\documentclass[10pt]{article}

\usepackage{PRIMEarxiv}

\usepackage[utf8]{inputenc} 
\usepackage[T1]{fontenc}    
\usepackage{hyperref}       
\usepackage{url}            
\usepackage{booktabs}       
\usepackage{amsfonts}       
\usepackage{nicefrac}       
\usepackage{enumitem}
\usepackage{lipsum}
\usepackage{multirow}
\usepackage[linesnumbered,ruled,vlined]{algorithm2e}
\SetKwInput{Input}{Input}   
\SetKwInput{Output}{Output} 
\usepackage{amsmath, amssymb, mathrsfs, bbm}
\usepackage{tabularx}
\usepackage{setspace}
\usepackage{subcaption}
\usepackage{xcolor}         
\usepackage{fancyhdr}       
\usepackage{graphicx}       
\usepackage[backend=biber,style=ieee,sorting=none,natbib=true]{biblatex}
\usepackage{authblk}

\graphicspath{{media/}}     
\usepackage{bm}

\title{Detection and Geographic Localization of Natural Objects in the Wild: A Case Study on Palms
}

\author[1,*]{Kangning Cui}
\author[2,*]{Rongkun Zhu}
\author[3]{Manqi Wang}
\author[1]{Wei Tang}
\author[3]{Gregory D. Larsen}
\author[4]{Victor P. Pauca}
\author[4]{Sarra Alqahtani}
\author[4]{Fan Yang}
\author[3]{David Segurado}
\author[5]{David Lutz}
\author[1]{Jean-Michel Morel}
\author[3]{Miles R. Silman}

\affil[1]{Department of Mathematics, City University of Hong Kong}
\affil[2]{Department of Computer Science, Xidian University, Xi'an, Shaanxi, China}
\affil[3]{Department of Biology, Wake Forest University, Winston-Salem, NC, USA}
\affil[4]{Department of Computer Science, Wake Forest University, Winston-Salem, NC, USA}
\affil[5]{School of Arts \& Sciences, Colby-Sawyer College, New London, NH, USA}

\begin{document}
\maketitle

\let\thefootnote\relax
\noindent\footnotetext{$*$ Kangning Cui and Rongkun Zhu contributed equally to this work. Corresponding author: kangnicui2-c@my.cityu.edu.hk}

\onehalfspacing
\begin{abstract}
Palms are ecologically and economically indicators of tropical forest health, biodiversity, and human impact that support local economies and global forest product supply chains. While palm detection in plantations is well-studied, efforts to map naturally occurring palms in dense forests remain limited by overlapping crowns, uneven shading, and heterogeneous landscapes. We develop PRISM (Processing, Inference, Segmentation, and Mapping), a flexible pipeline for detecting and localizing palms in dense tropical forests using large orthomosaic images. Orthomosaics are created from thousands of aerial images and spanning several to hundreds of gigabytes. Our contributions are threefold. First, we construct a large UAV-derived orthomosaic dataset collected across 21 ecologically diverse sites in western Ecuador, annotated with 8,830 bounding boxes and 5,026 palm center points. Second, we evaluate multiple state-of-the-art object detectors based on efficiency and performance, integrating zero-shot SAM~2 as the segmentation backbone, and refining the results for precise geographic mapping. Third, we apply calibration methods to align confidence scores with IoU and explore saliency maps for feature explainability. Though optimized for palms, PRISM is adaptable for identifying other natural objects, such as eastern white pines. Future work will explore transfer learning for lower-resolution datasets (0.5–1m).

\end{abstract}

\section{Introduction}

Palms (family Arecaceae) are vital to tropical ecosystems, serving as essential resources for pollinators and frugivores and influencing the evolution of dependent fauna~\cite{eiserhardt2011geographical,zambrana2007diversity}. Beyond their ecological roles, palms are deeply integrated in the livelihoods of rural and indigenous communities, providing food, construction materials, fuel, and medicine while supporting sustainable non-timber forest product markets. Particularly in regions like the Amazon, palms support subsistence practices and enhance resilience to socio-economic and environmental changes, reflecting the intricate links between biodiversity, human well-being, and sustainability~\cite{pitman2014distribution,malhi2014tropical,van2019palm,terborgh1986community}. Their distinctive star-shaped crowns make them well-suited for automated mapping using UAV imagery, supporting key efforts in biodiversity monitoring~\cite{sutherland2013identification,wagner2020regional}. 

\begin{figure}
    \centering
    \includegraphics[width=0.95\linewidth]{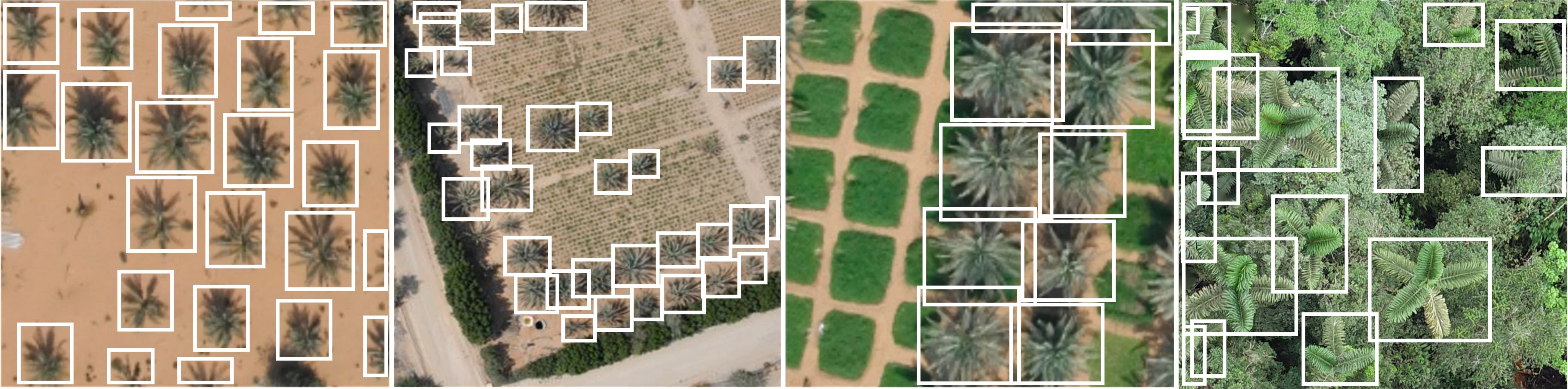}
    \caption{Palm Distribution Comparison. The first three images from previous studies~\protect\cite{gibril2021deep,jintasuttisak2022deep} feature evenly spaced palms or clear backgrounds, while the last represents our case with natural spacing, occlusions, and complex backgrounds in tropical forests.}
    \label{fig:comparebbox}
\end{figure}

Orthomosaic images are a valuable tool in remote sensing, with applications across various fields such as construction site monitoring, agricultural planning, environmental impact assessments, property development, and surveying, where accurate spatial data is essential~\cite{kucharczyk2021remote}. These images are created by stitching together hundreds or thousands of geolocated aerial frames into spatially coherent maps, enabling consistent analysis across landscapes. By providing a top-down view of large geographical areas, orthomosaics allow for precise measurement and detailed analysis of land features. Unlike standard aerial imagery, which contains perspective distortions, orthomosaics offer an in-scale representation of the ground. Low-flying drones enable efficient data collection without the need for extensive ground surveying~\cite{olson2021review}.

However, orthomosaics are not error-free and can be challenging to process. They often contain noise, artifacts, and stitching errors due to limited camera perspectives, variable lighting, wind-induced movement, and dynamic cloud cover during image acquisition~\cite{zhang2023aerial}. Additionally, resolution loss is a frequent byproduct of the stitching process. Typically stored in GeoTIFF format, orthomosaics can be exceptionally large, particularly in biodiversity conservation, forest health, and sustainable management applications, where a single file may range from 1 to 200 GB. As a result, image analysis must be performed in small patches that fit within the memory constraints of processing hardware.

Object-based detection techniques, widely used in remote sensing, are well-established in computer vision. However, detecting and localizing naturally occurring objects in orthomosaics is inherently challenging~\cite{tagle2019identifying,cui2024palmprobnet,qin2021ag,jintasuttisak2022deep}. Unlike structured plantations, where trees are arranged in grid patterns with ample spacing (see Figure~\ref{fig:comparebbox}), palms in tropical forests grow irregularly, often obscured by dense vegetation that distorts key diagnostic features. The distribution of palm species is highly imbalanced, with only a few dominant species present in sufficient abundance to train reliable detection models~\cite{li2016deep,freudenberg2019large}.
UAV-based imagery also introduces substantial lighting variability, as shadows from thick canopies and shifting sun angles create significant fluctuations in brightness and contrast across detection targets, further complicating model training and evaluation~\cite{gibril2021deep}. These challenges are amplified by the lack of high-quality labeled datasets, whose creation often requires extensive fieldwork and expert annotation~\cite{muscarella2020global,hidalgo2022sustainable}.

Lastly, manipulation of geographical information data requires domain expertise that is not readily available to computational scientists. A flexible pipeline that can process spatially referenced survey imagery and provide precise geographic target coordinates would support valuable downstream applications, leveraging state-of-the-art machine learning methods. These applications include monitoring palm abundance and distribution over time and across vast areas, and developing adaptive and modular survey methods. Additionally, balancing accuracy with interpretability and calibration is essential for trustworthy ecological decision-making~\cite{selvaraju2017grad,kuzucu2025calibration}. This study therefore introduces a novel end-to-end pipeline that addresses a major need in environmental monitoring, is computationally efficient and reliable, and can support variable-resource processing in field applications.

Our work presents the following main contributions:
\begin{enumerate}
    \item We \textbf{construct} and provide the  \textbf{PA}lm \textbf{L}ocalization in \textbf{M}ulti-\textbf{S}cale \textbf{(PALMS) dataset}. PALMS contains data from extensive fieldwork across 21 sites in western Ecuador, spanning a rainfall gradient from the Choco’s wettest forests to the dry edge of the Sechura desert, which captures corresponding gradients in palm species composition and canopy characteristics. For training, we annotated 1,500 image patches from 2 sites with 8,830 bounding boxes. For validation, we manually marked 5,026 palm crown centers from 4 reserves to direct compare detected georeferenced centers with ground truth.

    \item We \textbf{develop the PRISM pipeline}, a unified and modular pipeline for natural object detection, segmentation, and counting. PRISM integrates object detection with zero-shot segmentation to generate georeferenced palm coordinates. The modular design allows easy interchange of detection/segmentation models, while calibration analysis and saliency maps enhance trustworthiness and interpretability for ecological applications.
    
    \item We apply PRISM to the PALMS dataset and \textbf{validate  generalization} across four reserves with diverse environmental characteristics and species compositions. PRISM accurately localizes palm centers with strong ground-truth alignment. Benchmarking results show high inference efficiency across diverse hardware settings, supporting large-scale ecological monitoring.
    
\end{enumerate}

\section{Related Work}
\label{sec:related}

\subsection{Palm Detection and Localization}

Advancements in UAV technology, image stitching, and machine learning have driven significant progress in palm detection, segmentation, and localization from orthomosaic imagery. However, most studies have focused on commercially valuable species, such as oil and date palms, given their economic importance ~\cite{li2016deep,gibril2021deep,zheng2021growing,jintasuttisak2022deep,putra2023automatic}. For instance,~\citeauthor{li2016deep}~\cite{li2016deep} used CNNs with a sliding window for oil palm counting in Malaysia, while~\citeauthor{gibril2021deep}~\cite{gibril2021deep} developed a U-Net variant for enhanced date palm segmentation in UAE. \citeauthor{zheng2021growing}~\cite{zheng2021growing} proposed a Faster R-CNN variant with refined feature extraction and a hybrid class-balanced loss to monitor individual oil palm growth. More recently, YOLO-based approaches have been adopted:~\citeauthor{jintasuttisak2022deep}~\cite{jintasuttisak2022deep} applied YOLOv5 for detecting date palms from UAV imagery over UAE farmlands, while~\citeauthor{putra2023automatic}~\cite{putra2023automatic} employed YOLOv3 to detect and count oil palm trees for sustainable agricultural monitoring in Indonesia.

In contrast, the detection and localization of naturally occurring palms in tropical forests is largely underexplored. \citeauthor{tagle2019identifying}~\cite{tagle2019identifying} pioneered palm crown identification using random forest, showing machine learning’s potential for individual palm counting. \citeauthor{ferreira2020individual}~\cite{ferreira2020individual} applied a fully convolutional neural network with morphological operations to refine palm species segmentation. \citeauthor{wagner2020regional}~\cite{wagner2020regional} leveraged U-Nets and very high-resolution (0.5 m) multispectral imagery from the GeoEye satellite to map canopy palms over a large region of the Amazon rainforest.

\subsection{Object Detection and Zero-Shot Segmentation}
\label{sec:object}

Object detection, a core computer vision task, identifies and localizes objects via bounding boxes~\cite{zou2023object} and underpins advanced applications such as image segmentation and object tracking~\cite{wang2022sygnet,ma2024rethinking,li2024CPDR}. The field is dominated by methods using You Only Look Once (YOLO) and Detection Transformer (DETR). 

The YOLO family frames detection as a regression task balancing speed and accuracy. These methods often generate overlapping detections, which are typically resolved by a handcrafted process known as non-maximum suppression (NMS). YOLOv8~\cite{yolov8} enhances detection through advanced backbone and neck architectures for feature fusion, and an anchor-free detection head optimized for accuracy and speed. YOLOv9~\cite{yolov9} introduces programmable gradient information and the generalized efficient layer aggregation network to address information loss. YOLOv10~\cite{yolov10} eliminates NMS through consistent dual assignments during training and one-to-one inference matching, coupled with a refined CSPNet backbone and a lightweight classification head to reduce computational cost. YOLO11~\cite{yolo11} further enhances performance with a refined CSP bottleneck, hybrid attention, and adaptive anchors with extended IoU loss.

DETR~\cite{carion2020end} directly predicts object sets using learned queries, bypassing the need for post-processing such as NMS. DINO~\cite{zhang2022dino} enhances DETR with contrastive denoising and hybrid query initialization, while DDQ-DETR~\cite{zhang2023dense} introduces dense query assignment for improved one-to-one inference matching. RT-DETR~\cite{zhao2024detrs} optimizes DETR for real-time use via a hybrid encoder and multi-scale feature fusion.


Segment Anything Models (SAMs) are advanced segmentation models capable of segmenting any object in images using prompts such as points, boxes, or text~\cite{kirillov2023segment,ravi2024sam,mobile_sam}. Trained on the SA-1B (1 billion masks, 11 million images), SAM enables zero-shot inference and often surpasses fine-tuned methods in accuracy and efficiency~\cite{kirillov2023segment}. Its architecture features a ViT for image encoding, a prompt encoder to process input prompts, and a mask decoder that fuses features from both to generate segmentation masks. SAM 2~\cite{ravi2024sam}, trained on the SA-V dataset (50.9k videos, 642.6k masks), enhances video segmentation and object tracking by refining multi-scale feature extraction. Mobile SAM~\cite{mobile_sam} optimizes SAM for mobile use by simplifying the image encoder and using decoupled distillation, enhancing speed without compromising segmentation quality.

\begin{figure}
    \centering
    \includegraphics[width=0.7\linewidth]{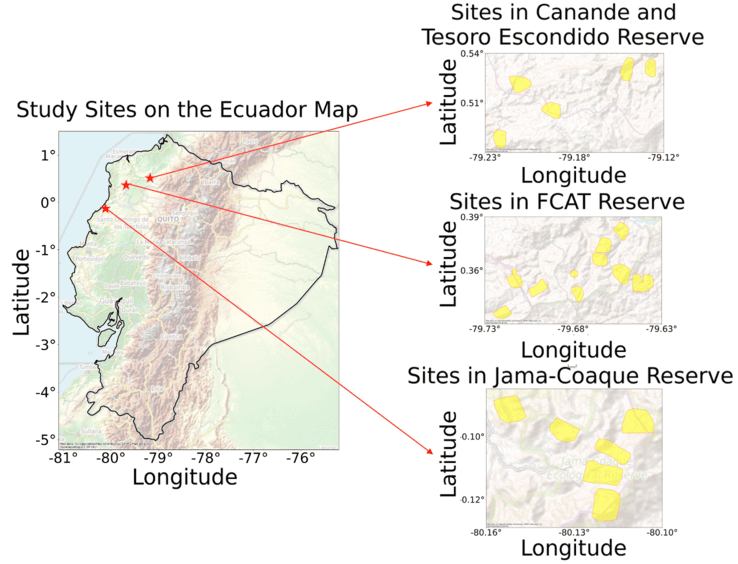}
    \caption{Geographic Locations of Study Sites. The left panel shows a map of Ecuador with red stars marking the study regions. The right panels zoom in on 21 study areas within four ecological sites.}
    \label{fig:map}
\end{figure}

\section{PALMS Dataset}
\label{sec:data}

In this section, we introduce the PALMS (\textbf{PA}lm \textbf{L}ocalization in \textbf{M}ulti-\textbf{S}cale) dataset. The following paragraphs detail the study sites, data collection, and manual labeling process.

\paragraph{Study Sites:} Data in this study (See Figure~\ref{fig:map}) come from western Ecuador's Choco tropical forest, including \textit{Fundación para la Conservación de los Andes Tropicales Reserve and adjacent Reserva Ecológica Mache-Chindul park} (FCAT; 00$^\circ$23'28'' N, 79$^\circ$41'05'' W), \textit{Jama-Coaque Ecological Reserve} (00$^\circ$06'57'' S, 80$^\circ$07'29'' W), \textit{Canande Reserve} (0$^\circ$31'34'' N 79$^\circ$12'47'' W), and \textit{Tesoro Escondido Reserve} (0$^\circ$33'16'' N 79$^\circ$10'31'' W). FCAT is a high diversity humid tropical forest at elevation $\sim$500m, receiving $\sim$3000 mm yr$^{-1}$ precipitation with persistent fog during drier period. Jama-Coaque ranges from the boundary of the tropical moist deciduous/tropical moist evergreen forest at the lower elevations ($\sim$1000 mm precipitation yr$^{-1}$, $\sim$250 m asl) to fog-inundated wet evergreen forests above 580m to 800m. Canande (350–500 m elevation) and Tesoro Escondido ($\sim$200 m elevation) are lowland everwet Choco forests, both receiving 4000–5000 mm yr$^{-1}$ precipitation with no month experiencing drought stress or precipitation below 100 mm. These forests host several palm species with exposed canopy crowns, including the economically important \textit{Iriartea deltoidea}, \textit{Socratea exorrhiza}, and \textit{Oenocarpus bataua}, with lesser amounts of \textit{Attalea colenda} and \textit{Astrocaryum standleyanum}, and species composition varying across study sites~\cite{browne2016diversity,lueder2022functional}.



\paragraph{Data Collection:} We collected UAV imagery in two stages, capturing 8,845 photos across 21 areas spanning 1,995 hectares, with a ground sampling distance under 6 cm. In June 2022, the first stage covered 95 hectares and produced 387 photos, while the second stage in February 2023 surveyed 1,900 hectares and captured 8,458 photos. Both missions used a DJI Phantom 4 RTK drone equipped with a 1'' CMOS sensor and GS RTK for mission planning. The first mission flew at 90 meters above ground level with 70\% sidelap and 80\% frontlap, while the second operated at 150 meters. The collected images were processed for subsequent analysis. To enable landscape-level forest analysis, we created orthomosaics and conducted post-processing steps using Agisoft Metashape 2.0, including noise removal, edge trimming, and the generation of digital surface and terrain models.

\paragraph{Manual Labels:} To enable fine-tuned palm detection and zero-shot segmentation, we curated a dataset of 1,500 images (800$\times$800 pixels) from two FCAT reserve sites, capturing varied quality and palm density typical of natural forests. Manual annotation of palm crowns and isolated leaves faced challenges from vegetation overlap and orthomosaic distortions. Three trained experts initially labeled bounding boxes, followed by iterative refinement using a YOLOv8 detector trained on these annotations to identify and correct labeling errors or omissions. For landscape-scale validation, a trained expert annotated 5,026 georeferenced palm centers on four orthomosaics using ArcGIS Pro 3.3.1, with predicted coordinates from PRISM to refine annotations. This hybrid human-model workflow addressed the inherent complexity of labeling in natural forest environments, ensuring robust annotations for both detection and geospatial validation.

\begin{figure*}[t]
    \centering
    \includegraphics[width=\linewidth]{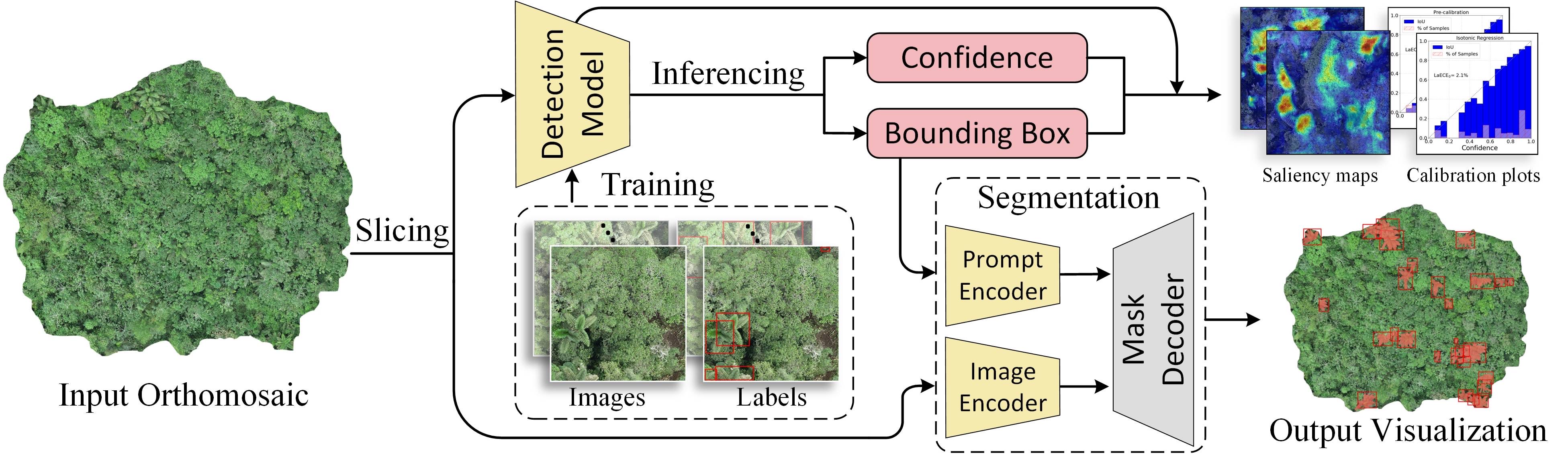}
    \caption{PRISM Pipeline Overview. The detection model, trained on the PALMS dataset, processes orthomosaic slices to generate confidence scores and bounding boxes. These bounding boxes are refined and serve as prompts, along with the sliced input images, for zero-shot segmentation. The bounding boxes and confidence scores are further utilized for saliency map generation and calibration analysis.}
    \label{fig:workflow}
\end{figure*}

\section{PRISM Pipeline}
\label{sec:method}

PRISM (Processing, Inference, Segmentation, and Mapping) is an end-to-end pipeline that process orthomosaic images to generate georeferenced palm coordinates alongside bounding boxes and segmentation masks for visualization. Fine-tuned on the PALMS dataset, PRISM addresses challenges specific to dense rainforest environments, such as irregular palm distributions and overlapping crowns. By integrating fine-tuned detection models with zero-shot segmentation, PRISM ensures adaptability to a wide range of environmental tasks. Figure~\ref{fig:workflow} illustrates the PRISM pipeline, with the details of each model component discussed in the following sections.

\paragraph{Detection Model:} We selected YOLOv10~\cite{yolov10} for its speed and performance. YOLOv10 introduces a consistent dual assignment strategy for NMS-free training, which combines one-to-many and one-to-one label assignments for enriched supervision while eliminating post-processing NMS. This approach uses a unified matching metric: \[m(\alpha, \beta) = s \cdot p^{\alpha} \cdot \text{IoU}(\hat{b}, b)^{\beta},\] where \( p \) is the classification score, \( \hat{b} \) and \( b \) are the predicted and ground truth (GT) bounding boxes, and \( s \) is the spatial prior, with \(\alpha\) and \(\beta\) balancing classification and IoU. Architecturally, YOLOv10 adopts an efficiency-accuracy-oriented design, integrating lightweight classification heads, spatial-channel decoupled downsampling, rank-guided block design, and advanced features such as large-kernel convolutions and partial self-attention, achieving robust performance with faster inference and fewer parameters. Trained on PALMS dataset, the model processes orthomosaic patches to output bounding boxes and confidence scores for further processing.

\paragraph{Segmentation Model:} We chose SAM 2~\cite{ravi2024sam} for zero-shot segmentation due to its superior segmentation quality and improved speed compared to its predecessor. SAM 2 uses a hierarchical image encoder for multi-scale feature extraction and an optimized architecture that reduces computational overhead while maintaining high precision. Its efficient prompt encoding and memory attention mechanism enable rapid mask refinement during inference. During inference, bounding boxes from the detection model undergo NMS to remove duplicates, as patches are cropped from the orthomosaic with a stride. The remaining bounding boxes and their surrounding image regions serve as input prompts for SAM 2 to generate segmentation masks. Final outputs include visualized bounding boxes and masks, alongside georeferenced coordinates derived from NMS-cleaned bounding box centers, which are subsequently used to quantify counting accuracy.

\begin{figure*}[t]
    \centering
    \includegraphics[width=\linewidth]{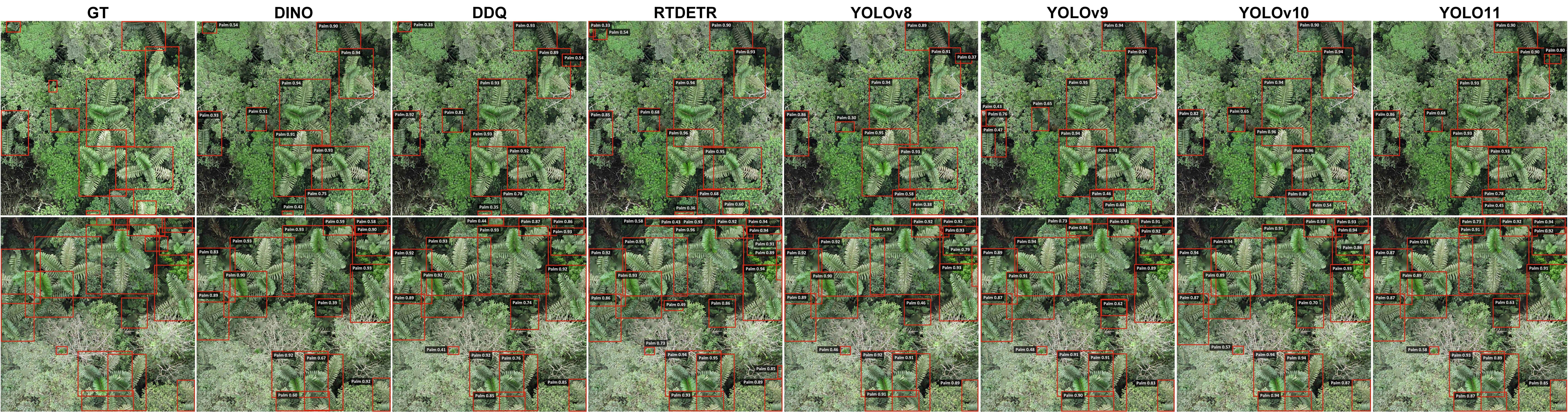}
    \caption{Comparison of Palm Detection Performance. Several models are compared in detecting palms, including small, occluded, and boundary-adjacent cases. All models perform well on large palms, even in occluded scenarios. DETR-based models excel at detecting small palms, while YOLO-based models perform better for partially visible palms on boundaries.}
    \label{fig:detection}
\end{figure*}

\begin{table*}[ht]
\centering
\caption{Comparative Analysis of Computational Efficiency and Detection Accuracy Among Models for PRISM. The table presents average performance metrics with standard deviations obtained from five random sampling experiments. FPS values were measured on an NVIDIA RTX 4090 24 GB GPU. Bold values indicate the best performance for each metric.}
\label{tab:detection}
\resizebox{\textwidth}{!}{%
\begin{tabular}{ccccccccc}
\toprule
\textbf{Model} & \textbf{GFLOPS} $\downarrow$ & \textbf{Params (M)} $\downarrow$ & \textbf{FPS} $\uparrow$ & \textbf{Precision} $\uparrow$ & \textbf{Recall} $\uparrow$ & \textbf{$\text{AP}_{50}$} $\uparrow$ & \textbf{$\text{AP}_{75}$} $\uparrow$ & \textbf{mAP} $\uparrow$ \\
\midrule
DINO    & 1920.3 & 218.2 & $18.98 \pm 0.95$ & $0.7629 \pm 0.0177$ & $0.8494 \pm 0.0071$ & $0.8169 \pm 0.0166$ & $0.5455 \pm 0.0150$ & $0.5102 \pm 0.0101$ \\
DDQ     & 1232.6 & 218.6 & $19.18 \pm 0.96$ & $0.7825 \pm 0.0124$ & $\mathbf{0.8566 \pm 0.0123}$ & $0.8541 \pm 0.0129$ & $0.6354 \pm 0.0137$ & $0.5736 \pm 0.0130$ \\
RT-DETR & 222.5 & 65.5 & $151.49 \pm 0.70$ & $\mathbf{0.8869 \pm 0.0230}$ & $0.7598 \pm 0.0310$ & $0.8416 \pm 0.0181$ & $0.6198 \pm 0.0181$ & $0.5769 \pm 0.0145$ \\
YOLOv8  & 226.7 & 61.6 & $174.92 \pm 0.86$ & $0.8729 \pm 0.0165$ & $0.7997 \pm 0.0203$ & $0.8667 \pm 0.0141$ & $0.6777 \pm 0.0137$ & $0.6148 \pm 0.0128$ \\
YOLOv9  & \textbf{169.5} & 53.2 & $114.96 \pm 0.30$ & $0.8763 \pm 0.0176$ & $0.7976 \pm 0.0209$ & $\mathbf{0.8741 \pm 0.0109}$ & $0.6762 \pm 0.0146$ & $0.6162 \pm 0.0122$ \\
YOLOv10 & 169.8 & \textbf{31.6} & $\mathbf{177.04 \pm 1.14}$ & $0.8716 \pm 0.0121$ & $0.7968 \pm 0.0089$ & $0.8626 \pm 0.0129$ & $\mathbf{0.6794 \pm 0.0112}$ & $\mathbf{0.6173 \pm 0.0090}$ \\
YOLO11 & 194.4 & 56.8 & $170.40 \pm 0.95$ & $0.8721 \pm 0.0095$ & $0.7896 \pm 0.0127$ & $0.8684 \pm 0.0108$ & $0.6677 \pm 0.0180$ & $0.6115 \pm 0.0109$ \\
\bottomrule
\end{tabular}
}
\end{table*}

\paragraph{Calibration:} Calibration analysis ensures model trustworthiness by aligning predicted confidences with empirical IoU, critical for ecological monitoring where overconfident false positives could mislead conservation decisions. We evaluate four calibration methods to quantify and mitigate errors in palm detection~\cite{kuzucu2025calibration}: (1) Linear regression (LR) maps logits to probabilities via a fitted linear function. (2) Isotonic regression (IR) fits a monotonic function by solving: $\min_{\hat{y}_1, \hat{y}_2, \dots, \hat{y}_n} \sum_{i=1}^n (y_i - \hat{y}_i)^2$, subject to $\hat{y}_i \leq \hat{y}_j, \forall i < j.$ (3) Temperature scaling (TS) divides logits by a learned temperature parameter \( T \) before applying a sigmoid. (4) Platt scaling (PS) applies logistic regression to map logits to probabilities as: \( p = \frac{1}{1 + \exp(-(a \cdot \text{logit} + b))} \).

\paragraph{Interpretability via Saliency Maps:} Saliency maps generated by Grad-CAM~\cite{selvaraju2017grad} enhance interpretability by visualizing how the model identifies palms, aiding ecological validation and debugging. For a class score $y$, Grad-CAM computes gradients of $y$ with respect to feature maps $A^k$, yielding weights  
$\alpha_k = \frac{1}{Z} \sum_{i,j} \frac{\partial y}{\partial A_{ij}^k}$, where \( Z \) is the number of spatial positions. The heatmap: $L_{\text{map}} = \text{ReLU}\left( \sum_{k} \alpha_k A^k \right)$ is then generated to highlight regions influencing predictions, such as palm crowns or diagnostic leaf patterns. This transparency helps ecologists verify whether the model focuses on biologically meaningful features (e.g., crown shapes) rather than spurious correlations (e.g., shadows).

\section{Experimental Results}
\label{sec:result}

This section evaluates the proposed pipeline across various tasks and study sites. We numerically compare the detection performance and computational efficiency of different models, and visually compare the results of different zero-shot segmentation models. Calibration metrics are examined to measure model trustworthiness by analyzing performance before and after calibration. Saliency maps are analyzed to trace attention shifts during inference. Next, we apply PRISM to orthomosaics with distribution shift to examine the counting performance in practical scenarios. Finally, real-time simulations are conducted to assess the detection and segmentation speed, as well as the computational demands across devices.

\begin{figure}[t]
    \centering
    \includegraphics[width=\linewidth]{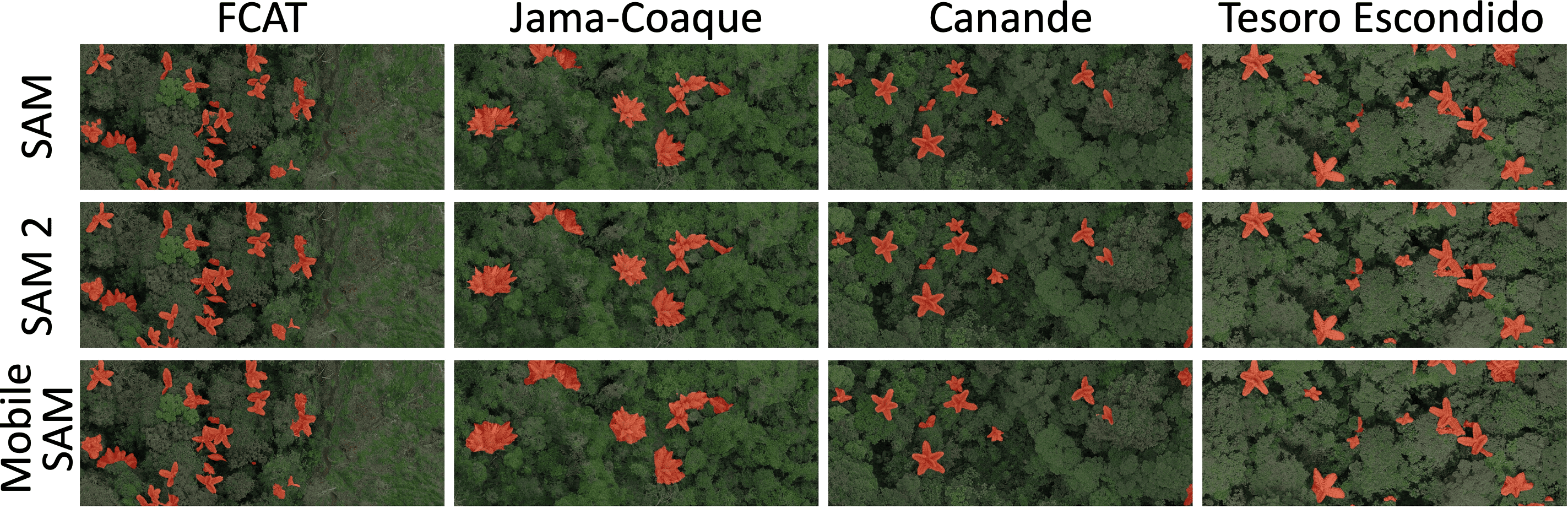}
    \caption{Zero-shot segmentation of SAM variants under distribution shifts. Rows correspond to SAM variants, while columns represent four distinct reserves. The box prompts were derived from the detection model trained on geographically distinct data.}
    \label{fig:segment}
\end{figure}

\begin{figure*}[t]
    \centering
    \begin{subfigure}[b]{0.11\linewidth}
        \centering
        \includegraphics[width=\textwidth]{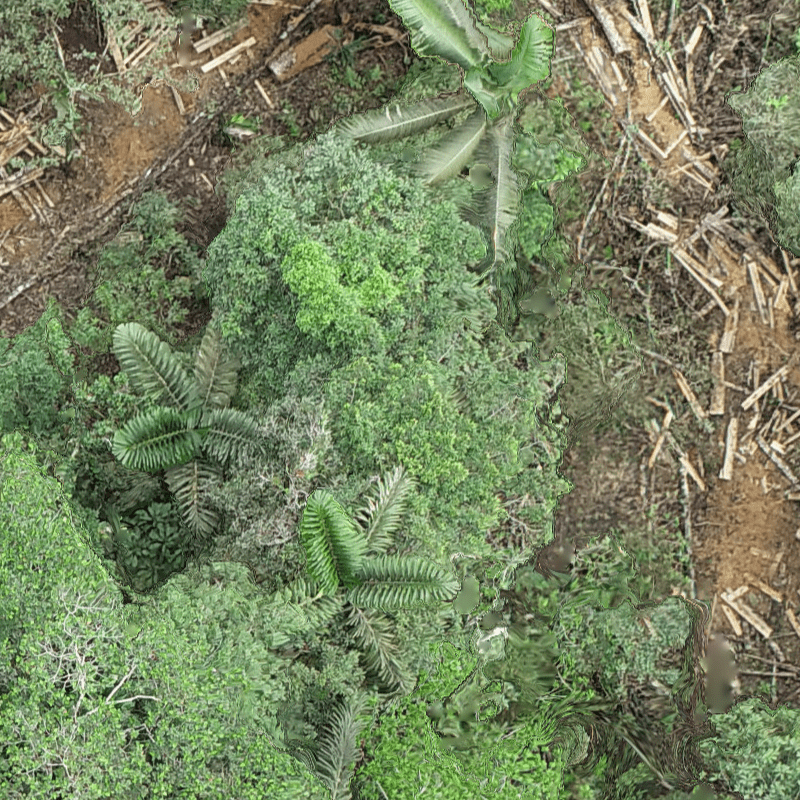}
    \end{subfigure}
    \hfill
    \begin{subfigure}[b]{0.11\linewidth}
        \centering
        \includegraphics[width=\textwidth]{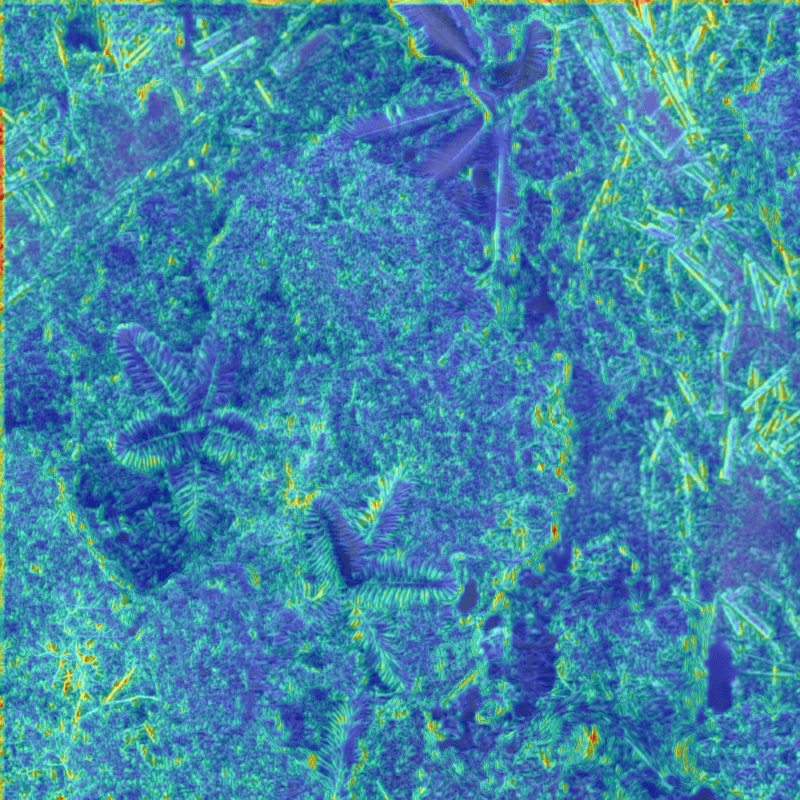}
    \end{subfigure}
    \hfill
    \begin{subfigure}[b]{0.11\linewidth}
        \centering
        \includegraphics[width=\textwidth]{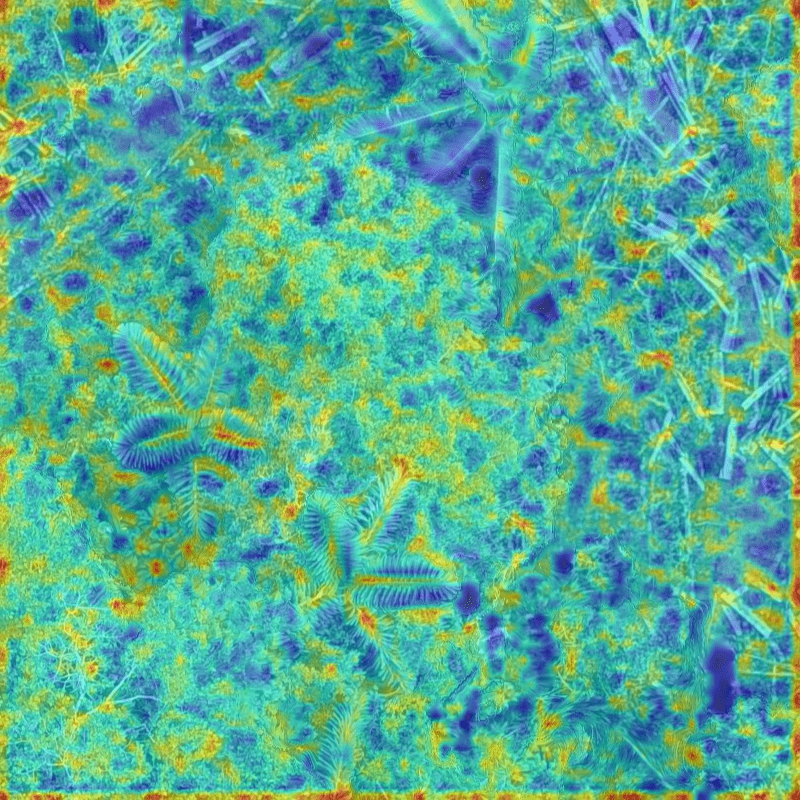}
    \end{subfigure}
    \hfill
    \begin{subfigure}[b]{0.11\linewidth}
        \centering
        \includegraphics[width=\textwidth]{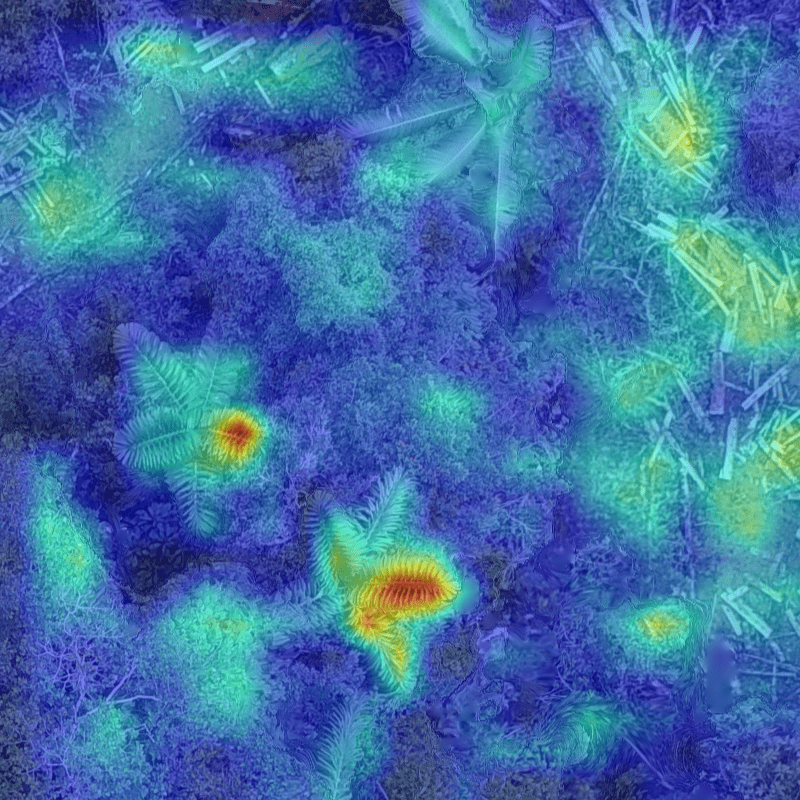}
    \end{subfigure}
    \hfill
    \begin{subfigure}[b]{0.11\linewidth}
        \centering
        \includegraphics[width=\textwidth]{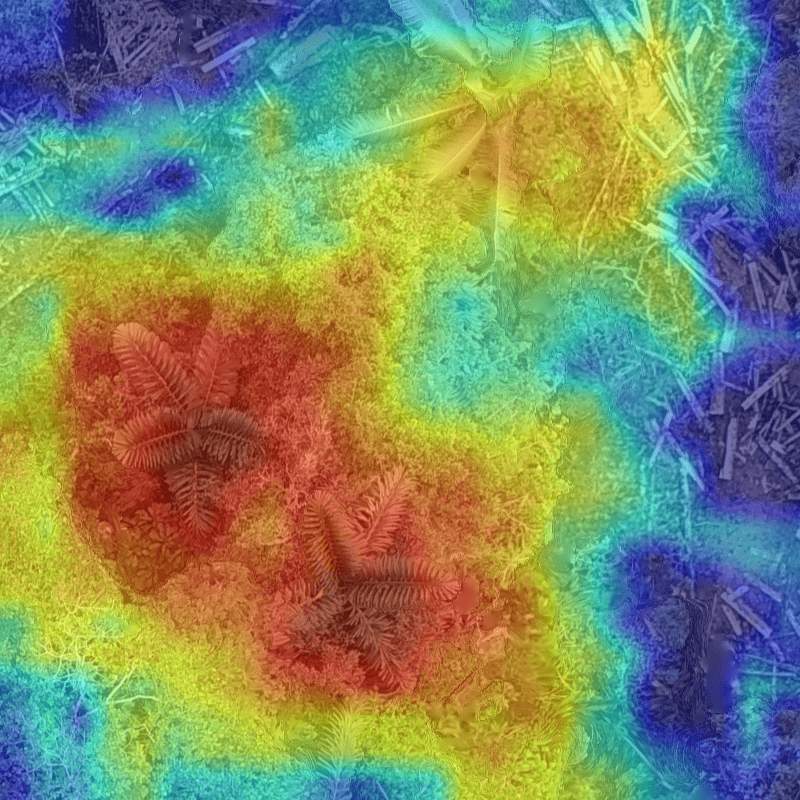}
    \end{subfigure}
    \hfill
    \begin{subfigure}[b]{0.11\linewidth}
        \centering
        \includegraphics[width=\textwidth]{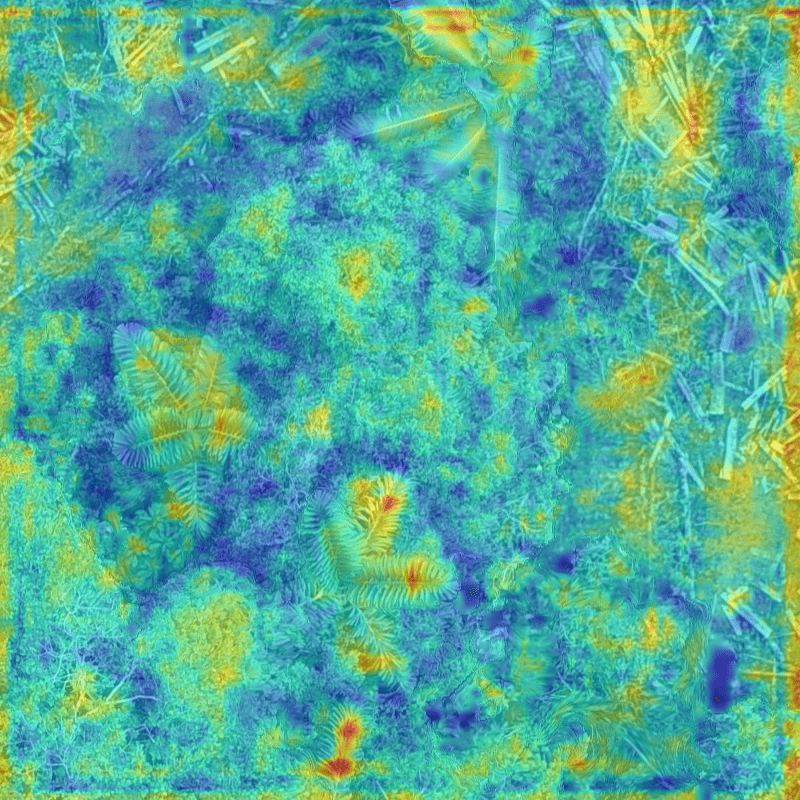}
    \end{subfigure}
    \hfill
    \begin{subfigure}[b]{0.11\linewidth}
        \centering
        \includegraphics[width=\textwidth]{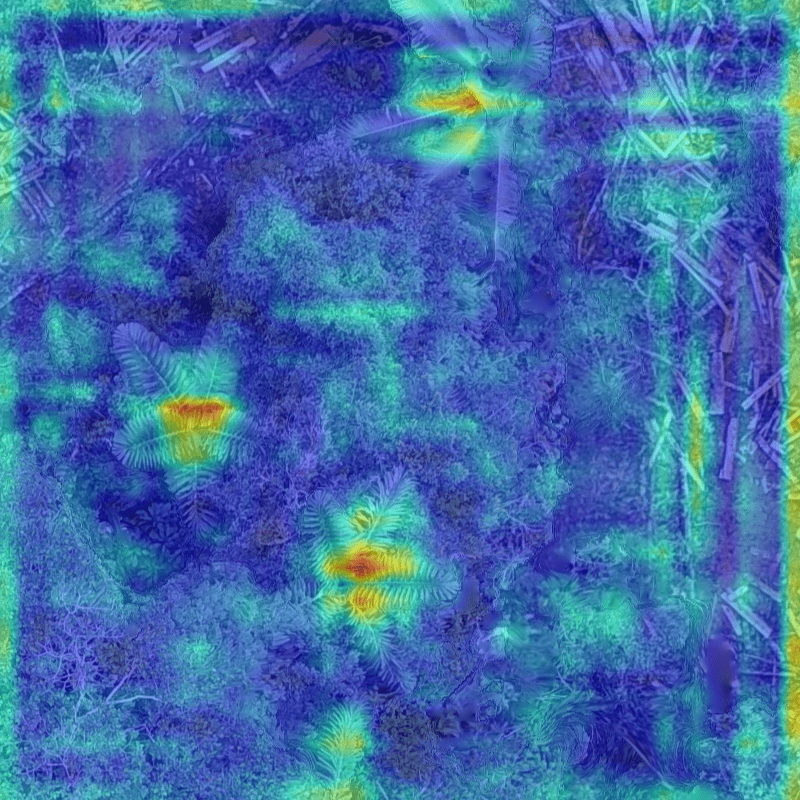}
    \end{subfigure}
    \hfill
    \begin{subfigure}[b]{0.11\linewidth}
        \centering
        \includegraphics[width=\textwidth]{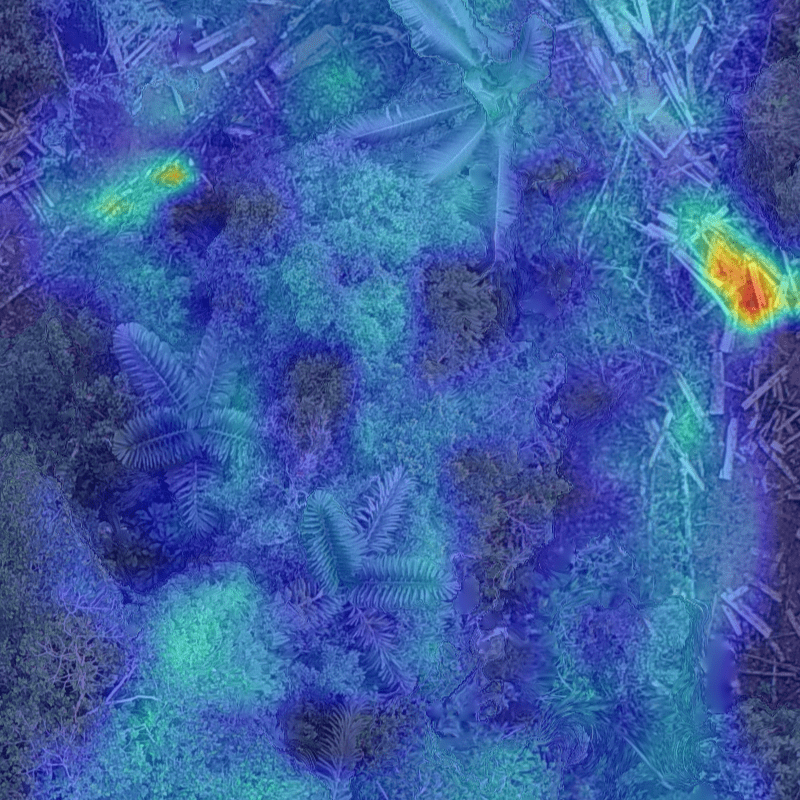}
    \end{subfigure}

    \begin{subfigure}[b]{0.11\linewidth}
        \centering
        \includegraphics[width=\textwidth]{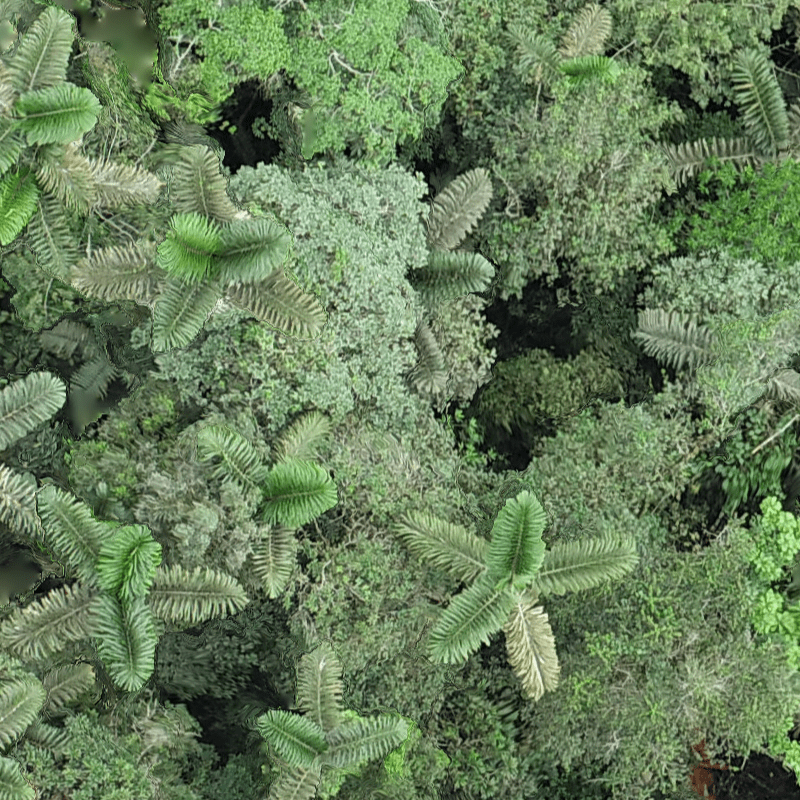}
        \subcaption{Input}
    \end{subfigure}
    \hfill
    \begin{subfigure}[b]{0.11\linewidth}
        \centering
        \includegraphics[width=\textwidth]{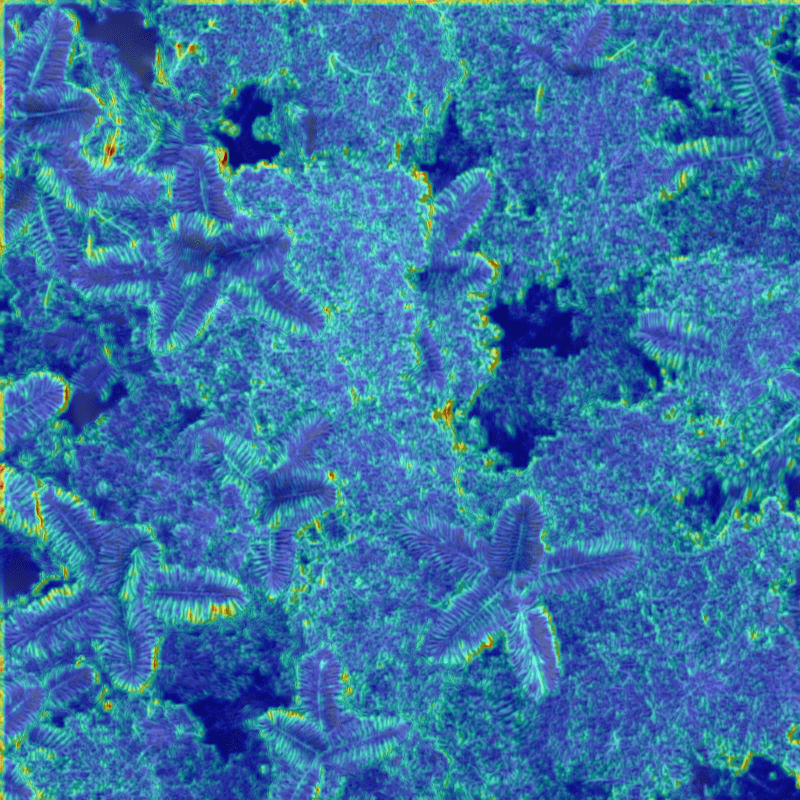}
        \subcaption{Layer 1}
    \end{subfigure}
    \hfill
    \begin{subfigure}[b]{0.11\linewidth}
        \centering
        \includegraphics[width=\textwidth]{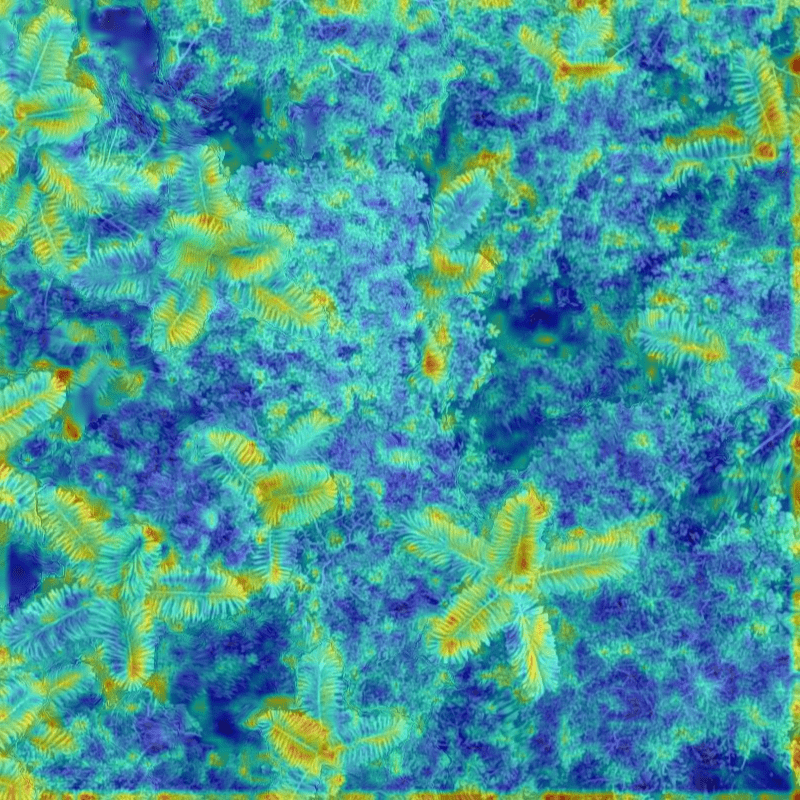}
        \subcaption{Layer 4}
    \end{subfigure}
    \hfill
    \begin{subfigure}[b]{0.11\linewidth}
        \centering
        \includegraphics[width=\textwidth]{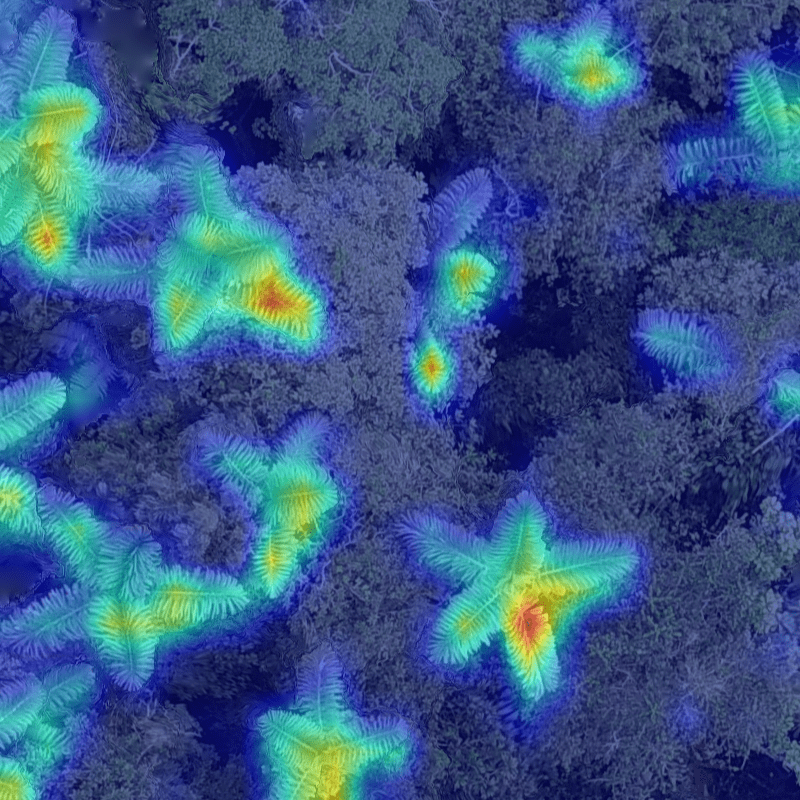}
        \subcaption{Layer 7}
    \end{subfigure}
    \hfill
    \begin{subfigure}[b]{0.11\linewidth}
        \centering
        \includegraphics[width=\textwidth]{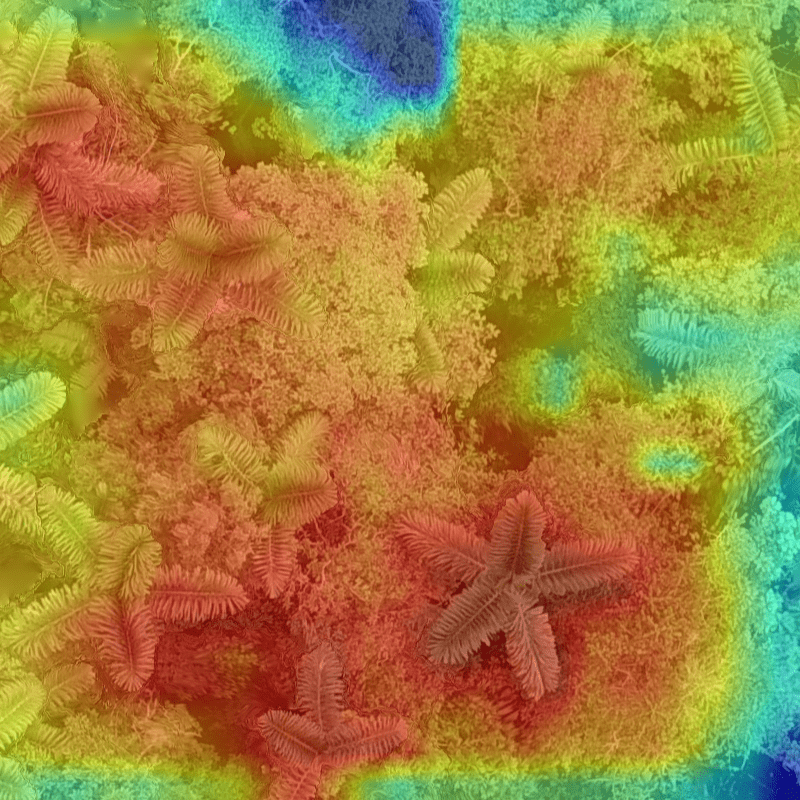}
        \subcaption{Layer 10}
    \end{subfigure}
    \hfill
    \begin{subfigure}[b]{0.11\linewidth}
        \centering
        \includegraphics[width=\textwidth]{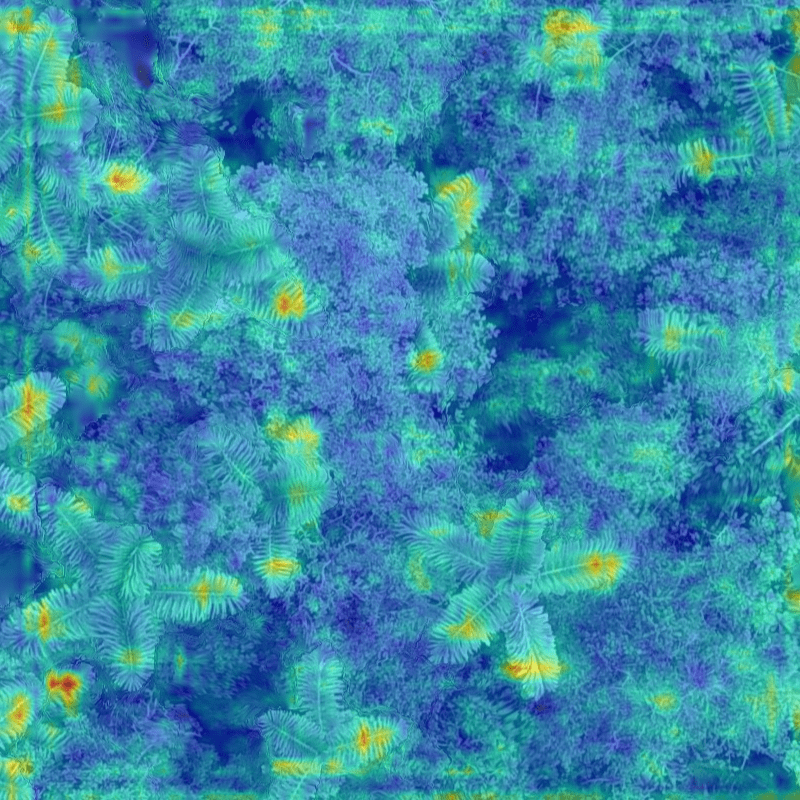}
        \subcaption{Layer 16}
    \end{subfigure}
    \hfill
    \begin{subfigure}[b]{0.11\linewidth}
        \centering
        \includegraphics[width=\textwidth]{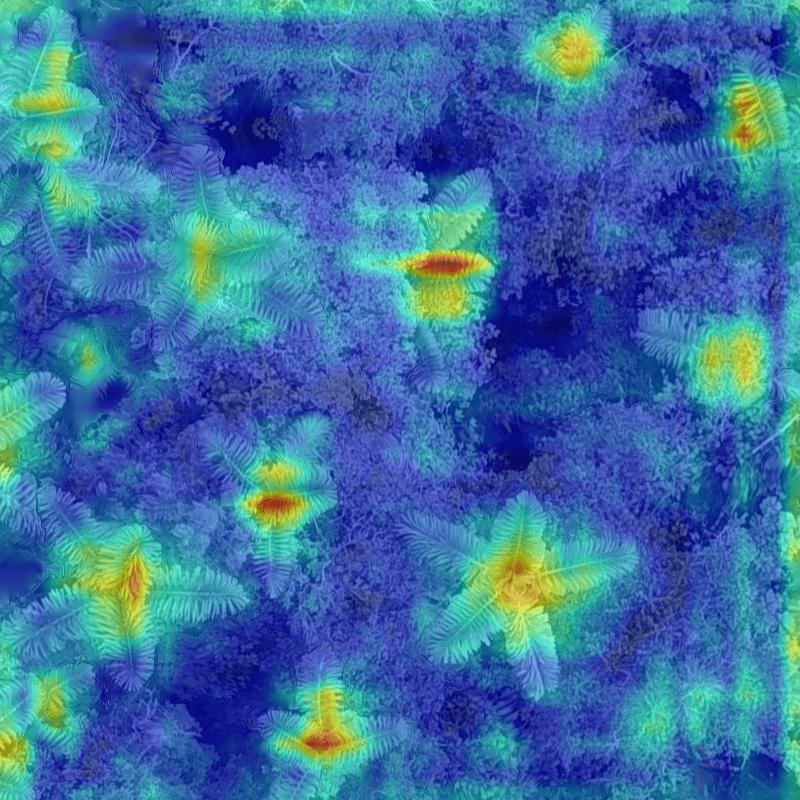}
        \subcaption{Layer 19}
    \end{subfigure}
    \hfill
    \begin{subfigure}[b]{0.11\linewidth}
        \centering
        \includegraphics[width=\textwidth]{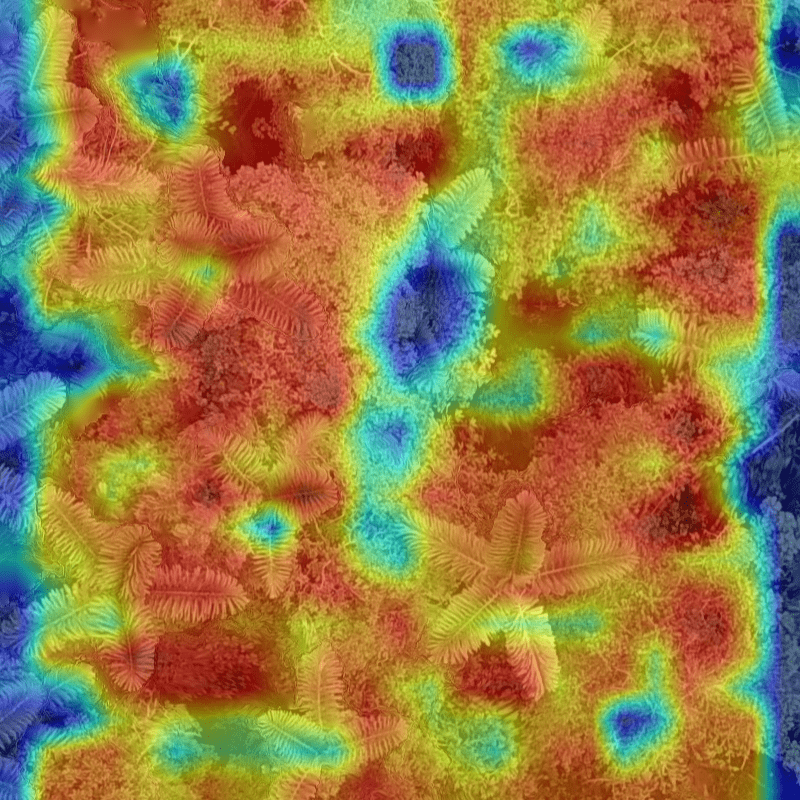}
        \subcaption{Layer 22}
    \end{subfigure}

    \caption{Feature Saliency in YOLOv10 via GradCAM Visualization. Early layers (b-c) extract boundary and edge details, mid-level layers (d-e) integrate spatial context, and deeper layers (f-h) specialize in scale-specific object detection.}
    \label{fig:saliency}
\end{figure*}

\subsection{Experimental Setting}

The detection models were trained on PALMS dataset with 1,500 images, split into 80\% for training, 10\% for validation, and 10\% for testing. Data splitting was randomly performed five times to ensure robust evaluation, with training and testing data drawn from different ecological sites, and validation data mixed from the two sites. YOLO variants and RT-DETR were trained for 100 epochs, while DINO and DDQ were trained for 30 epochs. Data augmentation techniques, including hue, saturation, and brightness adjustments, along with rotations, scaling, translations, and flips, were employed to enhance model robustness. For inference on landscape orthomosaics, a stride of 400 was used to ensure complete coverage, and NMS was applied to eliminate duplicate predictions. Training  was conducted on an RTX 4090 GPU, and testing was performed on multiple hardware platforms: RTX 3060 Laptop, RTX 4090 and H800 GPUs.

Detection performance was evaluated using precision, recall, average precision at IoU thresholds of 0.5 ($\text{AP}_{50}$), 0.75 ($\text{AP}_{75}$), and mean average precision (mAP) across multiple IoU thresholds. Consistency was assessed by reporting the standard deviation of these metrics across five random experiments. Computational efficiency was evaluated using GFLOPS and parameter counts (Params) to measure model complexity, and FPS to assess inference speed, illustrating their suitability for real-world applications. 

We evaluated the calibration performance of detectors by  
\[
\text{LaECE}_0 = \sum_{j=1}^{J} \frac{D_j\left| p_j - \text{IoU}_j \right|}{D}, \text{LaACE}_0 = \sum_{i=1}^{D} \frac{\left| p_i - \text{IoU}_i \right|}{D},
\] 
where \( J = 25 \) denotes the number of confidence bins, \( D \) and \( D_j \) represent the total detections and detections in bin \( j \), and \( p_j \) and \( \text{IoU}_j \) are the average confidence and IoU within bin \( j \). For \(\text{LaACE}_0\), \( p_i \) and \( \text{IoU}_i \) correspond to per-detection confidence and IoU~\cite{kuzucu2025calibration}. We evaluated the counting performance using retrieval ratios between predicted and GT centers within $d=5$ meters and report the median shift of successful matches to quantify spatial deviation. For real-time analysis, we assess the inference speed on raw images that were directly taken from UAVs.

\subsection{Detection and Segmentation Performance}

Table~\ref{tab:detection} compares the detection performance when training and testing on different geographical regions within the FCAT reserve, introducing natural distribution shifts. YOLOv10 demonstrates superior efficiency with 177.04 FPS and 31.6M parameters while achieving competitive accuracy, particularly in higher-quality metrics like $\text{AP}{75}$ (67.94\%) and mAP (61.73\%). It marginally outperforms other YOLO variants by 0.17\% AP$_{75}$ and 0.11\% mAP. DINO's lower AP$_{50}$ suggests reduced detection capability under shifted distributions. DDQ attains the best recall (85.66\%) and a solid $\text{AP}_{50}$, but its lower FPS limits real-time applicability. RT-DETR achieves the highest precision (88.69\%) but low recall (75.98\%), indicating numerous missed palms. These results position YOLOv10 as optimal for speed-critical deployments, with DDQ reserved for high-recall scenarios.

\begin{table}[t]
    \centering
    \caption{Comparison of Calibration Performance. The best calibration for each line is highlighted in bold. IR performs best LaECE$_0$, while performance for LaACE$_0$ varies by model.}
    \label{tab:calibraton}
    \resizebox{0.6\linewidth}{!}{%
    \begin{tabular}{ccccccc}
        \toprule
        \textbf{Metric}& \textbf{Model} & \textbf{Uncalibrated} & \textbf{IR} & \textbf{LR} & \textbf{PS} & \textbf{TS} \\ 
        \midrule
        \multirow{7}{*}{\textbf{LaECE$_0$}} 
            & DINO & 4.10\% & \textbf{2.90\%} & 3.50\% & 4.00\% & 3.70\% \\ 
            & DDQ & 2.40\% & 2.30\% & \textbf{1.60\%} & 1.80\% & 1.90\% \\ 
            & RTDETR & 8.40\% & \textbf{3.50\%} & 4.10\% & 3.60\% & 5.90\% \\ 
            & YOLOv8 & 3.60\% & \textbf{3.10\%} & 4.90\% & 4.30\% & 4.00\% \\ 
            & YOLOv9 & 3.30\% & \textbf{2.40\%} & 2.90\% & 2.70\% & 3.20\% \\ 
            & YOLOv10 & 4.30\% & 4.70\% & 4.40\% & 4.20\% & \textbf{3.60\%} \\ 
            & YOLO11 & 3.20\% & \textbf{2.20\%} & 3.80\% & 3.40\% & 3.50\% \\ 
        \midrule
        \multirow{7}{*}{$\textbf{LaACE$_0$}$} 
            & DINO & 0.16\% & 0.27\% & 0.15\% & \textbf{0.12\%} & 0.77\% \\ 
            & DDQ & 0.92\% & 0.61\% & \textbf{0.41\%} & 0.42\% & 0.55\% \\ 
            & RTDETR & 8.43\% & 2.95\% & 2.84\% & 2.80\% & \textbf{1.84\%} \\ 
            & YOLOv8 & \textbf{0.67\%} & 0.95\% & 0.92\% & 0.95\% & 1.41\% \\ 
            & YOLOv9 & \textbf{0.15\%} & 0.19\% & 0.42\% & 0.44\% & 1.01\% \\ 
            & YOLOv10 & 3.97\% & 4.12\% & 4.06\% & 4.17\% & \textbf{0.02\%} \\ 
            & YOLO11 & 0.78\% & 0.40\% & \textbf{0.02\%} & 0.06\% & 1.11\% \\ 
        \bottomrule
    \end{tabular}}
\end{table}

Detection analysis under challenging conditions reveals model-specific capabilities (see Figure~\ref{fig:detection}). All models detect large palms reliably, even when overlapping, but exhibit lower confidence on small palms. DETR-based methods demonstrate superior small-object detection performance (first-row top-left/bottom, second-row center), whereas YOLO-based methods excel in occluded scenarios with partial leaf visibility. In such cases, YOLO variants successfully detect individual palms while DETR-based approaches frequently merge detections (first-row bottom, second-row top-right). The sliding-window processing of orthomosaics in PRISM ensures that partially visible palms near edges are typically captured in adjacent patches, thereby compensating for missed edge detections.

We evaluated zero-shot segmentation performance using images from four reserves with distinct ecosystems, where bounding boxes generated by the detection model (trained on geographically distinct regions) introduce distribution shifts when used as prompts during inference, partly due to variations in palm species across sites. Figure~\ref{fig:segment} demonstrates robust zero-shot segmentation performance despite occasional errors from imperfect bounding box predictions. SAM occasionally produces fragmented palm leaves due to partial segmentation and background misclassification, while Mobile SAM exhibits background over-inclusion. SAM 2 offers more balanced results, handling areas near palm boundaries more effectively. Considering both segmentation quality and computational efficiency (analyzed in §5.5), SAM 2 is the optimal choice for the segmentation backbone in PRISM.

\begin{figure}[t]
    \centering
    \begin{subfigure}[b]{0.32\linewidth}
        \centering
        \includegraphics[width=\textwidth]{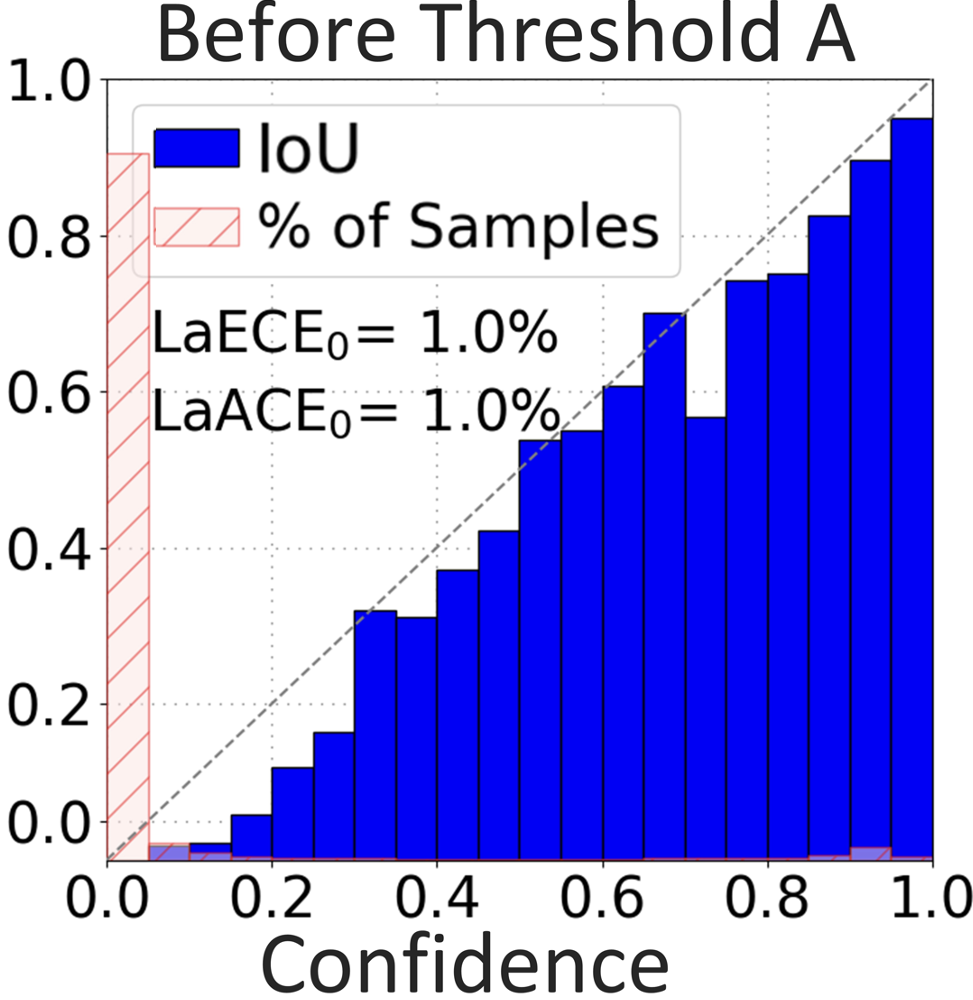}
    \end{subfigure}
    \hfill
    \begin{subfigure}[b]{0.32\linewidth}
        \centering
        \includegraphics[width=\textwidth]{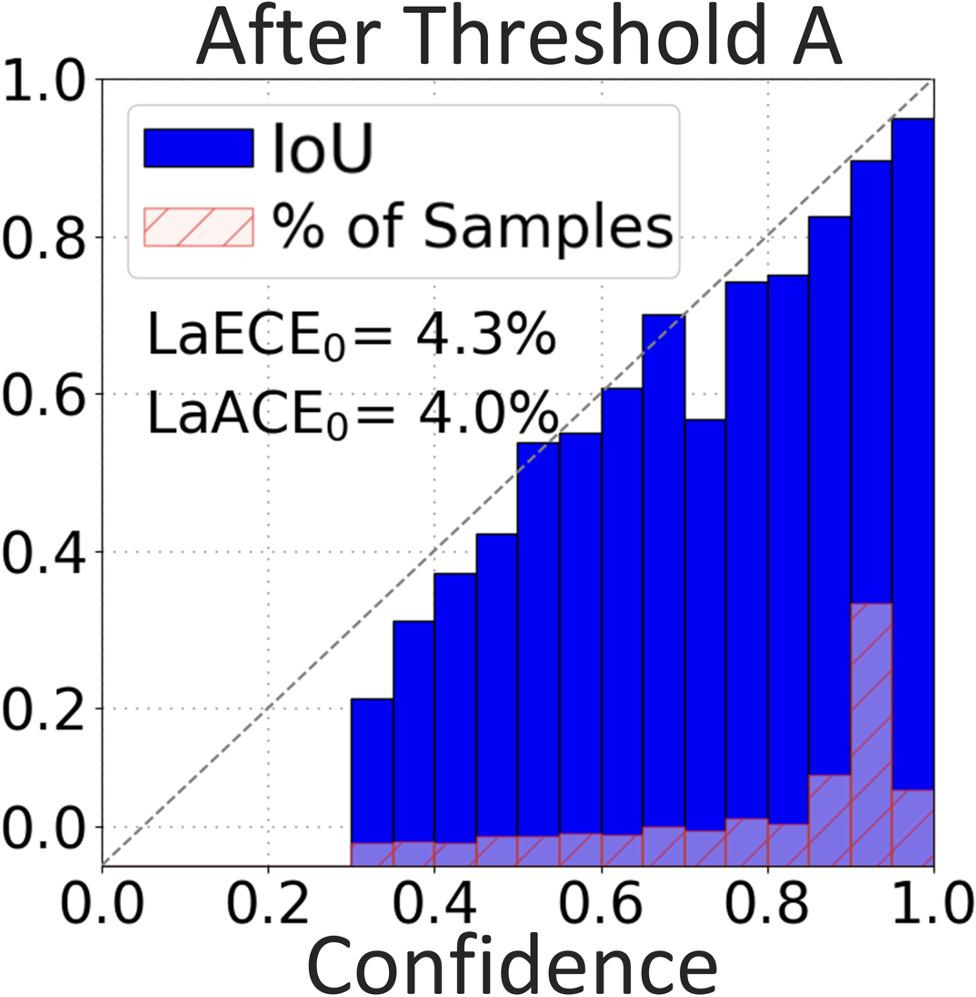}
    \end{subfigure}
    \hfill
    \begin{subfigure}[b]{0.32\linewidth}
        \centering
        \includegraphics[width=\textwidth]{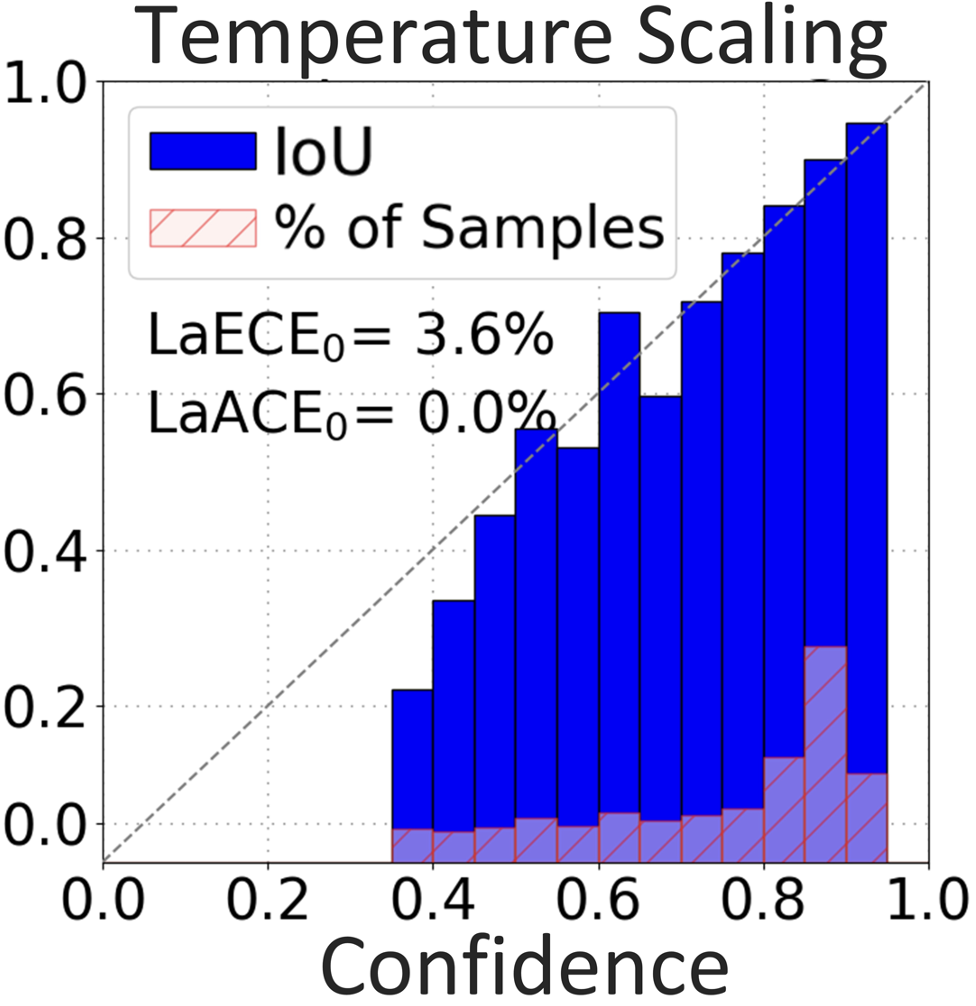}
    \end{subfigure}
    \caption{Impact of Thresholding and Calibration on YOLOv10’s Confidence Calibration. Calibration plots show confidence versus mean IoU, with LaECE$_0$ and LaACE$_0$ marked.}
    \label{fig:yolo10}
\end{figure}

\subsection{Calibration Analysis and Saliency Maps}

To enhance confidence reliability in palm detection, we evaluated calibration performance before and after applying calibration techniques. Confidence scores should align with IoU, and LRP-based thresholding ($A$) is used to filter out low-confidence detections~\cite{kuzucu2025calibration}. As shown in Table~\ref{tab:calibraton}, IR achieves the best LaECE$_0$ for most models, while YOLOv10 benefits most from TS and DDQ from LR. For LaACE$_0$, the optimal method varies -- TS improves RTDETR and YOLOv10, whereas LR is most effective for DDQ and YOLO11. Notably, no single method is universally best, and in some cases, lowering LaECE$_0$ increases LaACE$_0$. Figure~\ref{fig:yolo10} further illustrates the effect of LRP-based thresholding and calibration on YOLOv10's confidence distribution relative to IoU. Initially, low-confidence predictions ($<0.05$) dominate, skewing calibration metrics. LRP filtering refines the distribution by removing unreliable detections. Post-calibration, both LaECE$_0$ and LaACE$_0$ further decrease, indicating improved alignment between confidence scores and IoUs.

Grad-CAM saliency maps illustrate YOLOv10's hierarchical feature extraction across layers. Early layers (Figure~\ref{fig:saliency}(b-c)) capture fine-grained details, such as palm leaf boundaries and edge contrasts, progressively refining features from edges to full palm structures. Mid-level layers (Figure~\ref{fig:saliency}(d-e)) expand spatial context, enhancing palm localization. Deeper layers specialize in object scales: layer 16 emphasizes small features like individual leaves, layer 19 focuses on medium-sized palm crowns, and layer 22, designed for larger palms, shows limited relevance due to their absence in the scene.

\begin{figure}[t]
    \centering
    \begin{subfigure}[b]{0.7\linewidth}
        \centering
        \includegraphics[width=\textwidth]{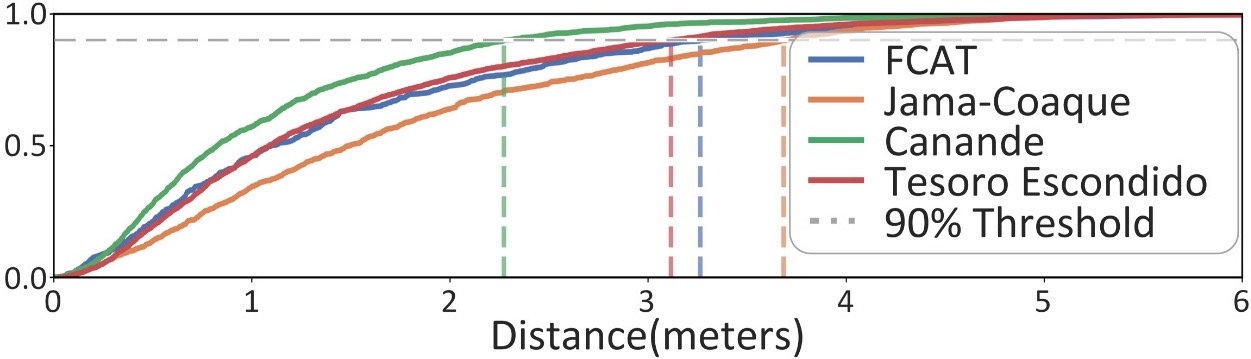}
        \subcaption{Pred2GT}
    \end{subfigure}
    
    \begin{subfigure}[b]{0.7\linewidth}
        \centering
        \includegraphics[width=\textwidth]{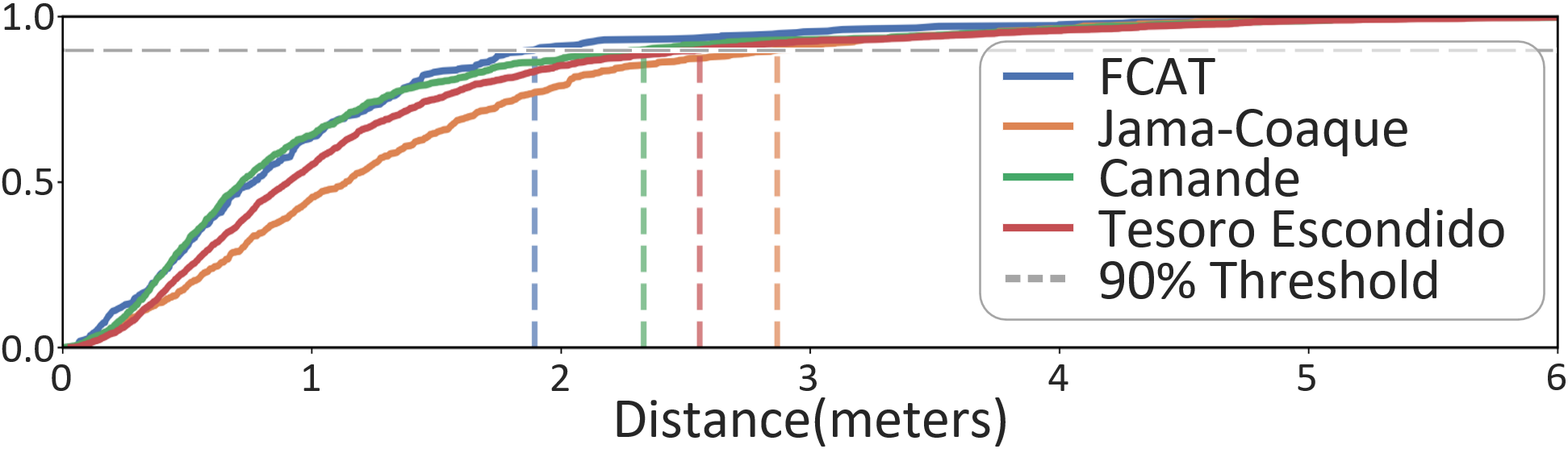}
        \subcaption{GT2Pred}
    \end{subfigure}
    \caption{Bidirectional Localization Shift Analysis: Cumulative distribution of Pred2GT and GT2Pred alignment. A 90\% cumulative threshold is indicated by the dashed line.}
    \label{fig:counting}
\end{figure}

\subsection{Counting Performance}

Table~\ref{tab:counting} reveals variations in palm counting robustness across sites. Pred2GT ratios (proportion of predictions matched to GT) remains consistently high (0.8956–0.9361), with FCAT achieving the highest retrieval rate (0.9361). In contrast, GT2Pred ratios (proportion of GT matched to predictions) vary markedly (0.7667–0.9253). Tesoro Escondido demonstrates strong bidirectional alignment with near-symmetric ratios, indicating balanced performance in localization and detection. FCAT, despite being a distinct sub-region, shows training-data familiarity  through its high GT2Pred ratio and the smallest GT2Pred median shift. However, its elevated Pred2GT shift (1.10m) suggests overfitting to site-specific features, yielding “precise but misplaced” predictions. Canande’s moderate Pred2GT ratio masks severe recall failures (GT2Pred: 0.767), where unmatched GT palms are likely true misses rather than localization errors. Jama-Coaque, with a low GT2Pred ratio (0.815) and the largest median shifts (Pred2GT: 1.50m, GT2Pred: 1.14m), reflects systematic challenges in both detection and localization. All sites maintain sub-1.5m median shifts, confirming robust localization despite occlusions or partial visibility. 

Figure~\ref{fig:counting} illustrates these trends: Canande’s smaller 90\textsuperscript{th} percentile Pred2GT shifts (Figure~\ref{fig:counting}a) align with its precise-but-conservative predictions, while FCAT and Tesoro Escondido’s dense canopies enable ``proximal counting'', where predictions align with clustered GT instances (e.g., overlapping crowns). This phenomenon partially inflates match rates in dense regions despite minor localization inaccuracies. FCAT’s strong GT2Pred performance highlights training benefits, whereas Jama-Coaque’s weaknesses underscore the need for targeted improvements in low-detection settings.

\begin{table}[t]
  \centering
  \caption{Counting performance across sites. Pred2GT and GT2Pred ratios quantify bidirectional alignment. Median distances indicate localization shifts.}
  \label{tab:counting}
  \resizebox{0.7\linewidth}{!}{
  \begin{tabular}{@{} l c c *{2}{c} *{2}{c} @{}}
    \toprule
    \multirow{2}{*}{\textbf{Site}} & 
    \multirow{2}{*}{\textbf{Area (ha)}} & 
    \multirow{2}{*}{\textbf{Counts}} & 
    \multicolumn{2}{c}{\textbf{Pred2GT}} & 
    \multicolumn{2}{c}{\textbf{GT2Pred}} \\
    \cmidrule(lr){4-5} \cmidrule(lr){6-7}
    & & & \textbf{Ratio} & \textbf{Median (m)} & \textbf{Ratio} & \textbf{Median (m)} \\
    \midrule
    \textbf{FCAT}             & 21.62  & 471  & 0.9361 & 1.10 & 0.8854 & 0.77 \\
    \textbf{Jama-Coaque}      & 111.93 & 952  & 0.9348 & 1.50 & 0.8151 & 1.14 \\
    \textbf{Canande}          & 101.20 & 1,273 & 0.8956 & 0.82 & 0.7667 & 0.72 \\
    \textbf{Tesoro Escondido} & 86.76  & 2,330 & 0.8981 & 1.09 & 0.9253 & 0.91 \\
    \bottomrule
  \end{tabular}%
  }
\end{table}

\begin{table}[t]
\centering
\caption{Inference time per image (seconds) across hardware configurations. Mean processing times ($\pm$ standard deviation) are computed over 20 raw images.}
\label{tab:speed}
\resizebox{0.65\linewidth}{!}{%
\begin{tabular}{@{}lcccc@{}}
\toprule
\textbf{GPU} & \textbf{YOLOv10} & \textbf{Mobile SAM} & \textbf{SAM} & \textbf{SAM 2} \\
\midrule
\textbf{RTX3060 Laptop} & 5.70$\pm$0.58 & 16.49$\pm$8.76 & 58.66$\pm$31.21 & 44.29$\pm$23.33 \\
\textbf{RTX4090} & 1.68$\pm$0.41 & 7.18$\pm$3.67 & 16.31$\pm$8.58 & 11.87$\pm$6.24 \\
\textbf{H800} & 1.21$\pm$0.37 & 5.92$\pm$3.00 & 14.05$\pm$7.27 & 10.33$\pm$5.31 \\
\bottomrule
\end{tabular}%
}
\end{table}

\subsection{Real-Time Simulation}  

We evaluated the computational feasibility for real-time onboard processing by simulating the detection-segmentation pipeline on raw UAV imagery using three GPU configurations as shown in Table~\ref{tab:speed}. Testing across 20 images with heterogeneous palm densities demonstrated robust detection performance: YOLOv10 achieves real-time inference speeds (1.21–5.70 s/image) with low temporal variance, confirming that mid-range hardware (an RTX3060 Laptop) remains capable of sustaining real-time detection tasks. Segmentation times, however, show substantial variability, as run-time scales linearly with detected palm count. Although segmentation provides valuable auxiliary visualization, its computational cost makes it optional for latency-critical deployments. These results confirm detection as the time-determining component, with YOLOv10's stability and speed meeting real-time UAV operational requirements.

\section{Conclusion}

We presented PRISM for automated palm detection and segmentation using UAV imagery, validated on the western Ecuador’s ecologically diverse reserves. PALMS dataset (21 sites, 8,830 bounding boxes, 5,026 georeferenced palm centers) captures climatic and species diversity critical for biodiversity monitoring. The modular pipeline achieves real-time processing across GPUs, with sub-1.5m median localization shifts under environmental distribution shifts, while calibration analysis and saliency maps ensure trustworthiness and interpretability. Future work will prioritize edge device deployment for UAV integration and field validation. PRISM’s design addresses challenges specific to palms -- occlusion, irregular spacing, and lighting variability -- which are common to detecting trees in structurally complex environments (e.g., \textit{eastern white pines}). Its robustness suggests adaptability to other ecologically critical tree species in wild ecosystems and lower-resolution satellite imagery (0.5–1 m), enhancing scalable ecological monitoring. 

\printbibliography

\end{document}